# Automatic extraction and 3D reconstruction of split wire from point cloud data based on improved DPC algorithm

Jia Cheng*

School of Computer Science and Technology, Tian Gong University, Tianjin, 300387, China

**Acknowledgement:**

Scientific research program of Tianjin Education Commission (NATURAL SCIENCE) （2019KJ017）

*Corresponding author:Jia Cheng

**Abstract**:In order to solve the problem of point cloud data splitting improved by DPC algorithm, a research on automatic separation and 3D reconstruction of point cloud data split lines is proposed. First, the relative coordinates of each point in the cloud point are calculated. Second, it is planned to develop a relative ensemble-based DPC swarm algorithm for analyzing the number of separation lines to determine all parts in the cloud content. Finally, fit each separator using the least squares method. iron. The cloud point of the resulting split subconductors has a clear demarcation line, and the distance between adjacent split subconductors is 0.45 m, divided by the four vertices of the square.

## 1 Introduction

In recent years, the generation of real 3D models based on a variety of materials has become an important part of obtaining product surface information, and has been widely used in computer graphics, computer vision, surveying and mapping, robotics, archaeology and other fields. . The raw data collected by search engines is usually represented by a 3D point cloud. Due to its simplicity and simplicity, it has gradually become a tool in a wide range of research and engineering applications. Using 3D digital technology to scan existing objects to obtain 3D surface data point cloud, and analyze and process the cloud content collection, and then perform accurate and fast 3D reconstruction, is the key issue of 3D digital visualization processing technology. 3D point cloud information processing technology has become the basis of current applications, and the development of 3D point cloud data processing theory and process has become the focus of current academic research. Image-based 3D reconstruction methods reconstruct by acquiring multiple 2D image sequences on the product surface. This paper deeply studies the key techniques of weather data processing in the reconstruction process, as shown in Figure 1.

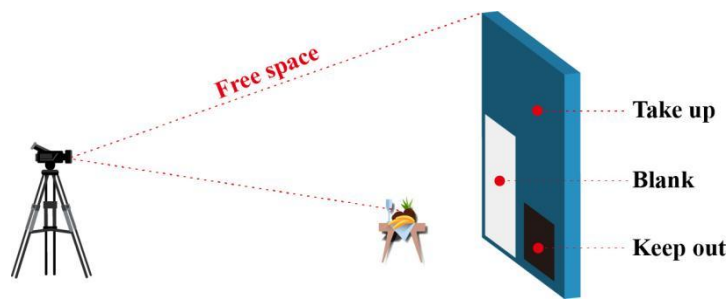

Figure 1Point cloud data split wire automatic extraction with 3D weight

## 2 Literature review

The 3D model reconstruction method based on point cloud can be measured from the execution efficiency and the accuracy of restoring the surface of the object. Therefore, it is particularly important to be able to construct a set of fast and quality reduction methods. In addition, another challenge in the processing of reconstruction algorithms is how to deal with 3D point sets of different scales. Since point cloud 3D reconstruction has important research value and practical value in various fields, it is necessary to ensure that the algorithm has certain applicability to different models when researching point cloud data reconstruction algorithms, such as in historical buildings. Digital files and terrain visualization require the storage of billions of incoming 3D points; real-time sensing systems generate millions of data points per second [1]. Large-scale point clouds make storage and subsequent processing inefficient. In addition, with the development of science and technology, 3D printing technology has begun to appear in people's field of vision. People often use 3D printing to create their favorite objects. The data point scale of these objects to be built is much smaller than other fields. It can better represent the geometric characteristics of the object to be measured.

Anne, B claimed the advantage of an air denoising method with multiple filters that preserves geometrically significant features [2]. Xu W. H. et al. Apply a mix of filters and fuzzy C-means groups to denoise the point cloud [3]. Lee, J. et al. Using the trilateral filtering function to filter and optimize the cloud data content, the algorithm will effectively preserve the features of the model surface. Sun et al. For example, restricting regions within regions to regions with similar vectors and changing the normal direction [4]. Modified, higher performance. Hai, H., et al. Using the R*-wood position index model, the weighted mean adaptive filtering method is applied to the scattered point cloud data, which is based on the local surface of the cloud point, but the size of the neighbors has a good influence on the filtering and smoothing effect [5]. Zhang, Y. Y. et al. Carry out multiple cloud computing sites by introducing professional concepts and improving ICP. The algorithm has many advantages in terms of registration. Li, X. et al. An accurate registration algorithm using the overlapping volume of the overlapping region as an error metric is required, with high registration accuracy and fast convergence [6]. Zhou, C. et al. Increase the curvature limit of this base to achieve automatic combination of cloud points [7]. Lu, S. H. et al. Utilize the process of international hierarchical block search to quickly register 3D scan data for adjacent local search terms. Chen Xijiang et al. Use the jet feature to find the same name points and get better registration [8].

On the basis of the current research, automatic extraction and 3D reconstruction of split traverse from point cloud data are proposed. The most basic of 3D reconstruction work is the acquisition of 3D spatial data, that is, the acquisition of position, texture and topology information. There are two traditional ways of data acquisition, one is the spatial coordinate acquisition method of discrete single point, and the other is the method of using two-dimensional image data to describe the relative position in space. For 3D models with complex structures, the data acquisition method of single point measurement is time-consuming and labor-intensive, and the efficiency is low. However, with the emergence of 3D laser scanners, the detailed features of the reconstructed object surface can be quickly acquired by scanning data acquisition, and the high precision can meet the needs of 3D reconstruction [9].

## 3 Based DPC Algorithm Improved Point Cloud Data Split Traverse Automatic Extraction

and 3D Reconstruction

**3.1 Improved 3D reconstruction method of point cloud data**

In this paper, a 3D point cloud reconstruction algorithm based on Delaunay triangulation is proposed. The purpose of this research is to create a manifold center for 3D point cloud data so that this surface can create the geometrical features of the object. The algorithm is improved, the cloud data is divided into three parts by using the local development area, and the method of dividing and overlapping the local triangulation map is adopted, and the gene content after segmentation is optimized. Before reconstructing the point cloud data, the point cloud data needs to be preprocessed. The goal is to establish an area around each blast source and set the initial behavior of each blast message to a non-destructive state. Distributing cloud data data using the development zone algorithm is an important part of algorithm development. For all data points in the classified data, determine all cloud data points in the k-order domain as a method. The necessary material is attached to it, the edge formed after joining is used as a boundary ring, and the boundary ring is further extended until the end is determined [10]. Delaunay triangulation of point cloud data is the key to algorithm improvement. It triangulates the data distribution cloud content using only the concepts of partitioning and overcoming. During distribution, use the map method as an instruction. The data is prepared for a plane that is locally triangulated using the 2D Delaunay triangulation algorithm. The next process usually involves the integration of the various installation sites and the subsequent optimization process. The operation diagram of this method is shown in Figure 2

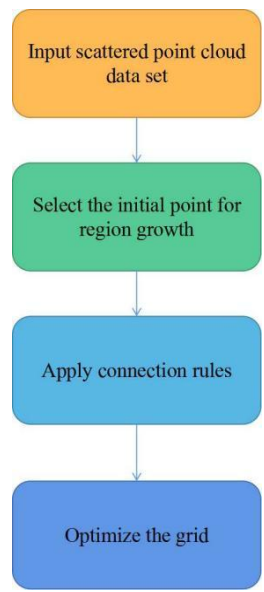

Figure 2 Program flow chart

**3.2 Current status of 3D reconstruction research**

At present, there are many methods of 3D reconstruction, which can be divided into the following types according to different data sources:

(1) Method based on traditional surveying and mapping results. This method takes the traditional surveying and mapping results as the base map, and completes the model reconstruction by editing the elevation and adding texture and other information in the 3D modeling software [11]. 3dsmax, CAD, Sketch Master, etc. are the most commonly used modeling software. However, although this method can obtain high model accuracy through human-computer interaction, the drawing cycle of the base map is too long and the cost is high, which has limitations in terrain modeling work.

(2) Procedures based on aerial imagery. The method uses mathematical methods to remove the contours of reconstructed objects from an image as source files or to directly construct the structure of the image. It is an important structural model in the field of photogrammetry. Large-scale 3D reconstruction of cities mainly uses the above-mentioned images and remote sensing images as data sources for this business. For example, Li Jinye et al. used segmentation to extract buildings and their shadows from remotely visible images. Its relationship takes the height of the building as the basic material for building construction. Tan Qulin et al. Models with better results from images with region segmentation, product distribution and performance. Tian Yi et al. The 3D model of Changzhou is obtained by applying the principle of oblique photography reconstruction, which is very effective [12].

### 3.3 Research status of point cloud data processing at home and abroad

Every year, top computer graphics conferences and journals such as Siggraph, Siggraph Asia, ACM TOG, and IEEETVCG publish a large number of papers related to point cloud processing. Conferences and journals are also constantly emerging with new related research work. In China, Zhejiang University is the first to study the method of point cloud processing; for the large-scale urban modeling technology of vehicle lidar, the Institute of Advanced Technology of the Chinese Academy of Sciences has also carried out a series of cutting-edge research; in addition, some other research institutions and universities such as Jilin University, Wuhan University, Shandong University, etc. are all carrying out research work on point cloud processing related technologies. At present, there are a lot of commercial software and open source software for point cloud processing. Commercial software such as Geomagic Studio, ImageWare, CopyCAD, RapidForm and other reverse engineering software; open source software includes Meshlab, Pointshop3D, Point CloudLibrary (PCL) and other open source platforms. The above commercial software and open source software integrate the latest technologies in many aspects, including noise reduction, segmentation, feature estimation, registration and reconstruction of point cloud data, etc. Powerful tool. In the current academic and industrial circles, the research on the key technologies of 3D point cloud data processing has attracted more and more attention, and it is also a hot issue in computer graphics and computer vision [13]. The processing technology of 3D point cloud data involved in this paper mainly includes: denoising and smoothing, initial registration, precise registration, and surface reconstruction.

### 3.4 Point cloud data acquisition technology

Traditional geometric modeling techniques such as computer-aided design, direct design of simple curves and surfaces to 3D models, and image-based modeling cannot accurately represent real-world objects with rich surface details, such as statues, trees, and so on. With the reduction of the cost of 3D scanning equipment and the improvement of software processing technology, the use of laser scanners and other acquisition equipment to collect the surface of objects and obtain realistic 3D digital models has gradually become one of the main modeling methods [14]. With the rapid development of 3D scanning technology and the increasing demand for 3D models in various fields, many different types of acquisition equipment have appeared. Granularity, accuracy, and quality of reconstructed models.

The discovery of triangulation data refers to the use of various modern means to record the surface data of the product, and to convert the field data into shared data of points, that is, to digitize the position of the object. [15]. In this way, the three-dimensional data received by the measuring equipment is very large, called "point cloud". In-depth research on the acquisition and expansion of existing heritage is the focus of many scholars. The rapid development of point cloud acquisition technology supports the use of 3D data in a variety of ways. Currently, the 3D technology industry can be divided into two categories, contact and non-contact, as shown in Figure 3.

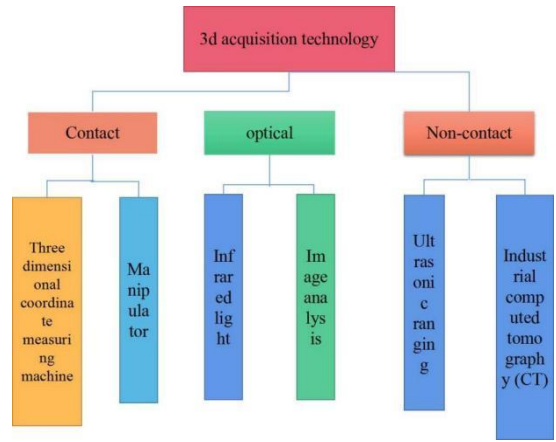

Figure 3 Point cloud data acquisition diagram

**4 Experiments and Research**

**4.1 Point cloud data processing**

surveying and mapping technology and the continuous updating of data acquisition equipment, the acquisition speed, anti-interference ability, ease of operation and real-time performance have been greatly improved. In particular, the emergence of unmanned aerial vehicles and 3D laser scanners occupies a dominant position in the acquisition of 3D space information. Figure 4 lists the cloud data volume of each station of the 3D laser scanner, the overall point cloud data volume of the UAV, and the collection and utilization time [16].

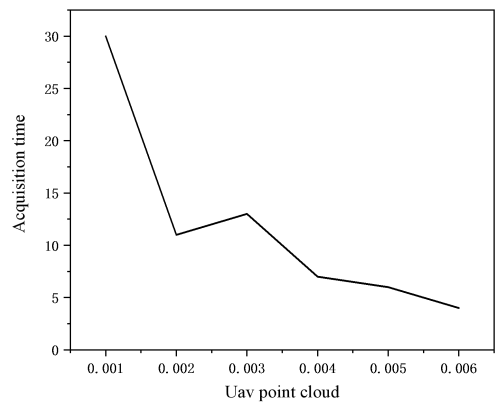

Figure 4 Point cloud data volume statistics

As can be seen from Figure 4, the total cloud content in this study area is 720 million, of which 22 million are received by drones; the cost of data collected from a single-station 3D laser scanner ranges from 20 million to 200 million large files. The amount of data is mixed with a large amount of redundant data, which will inevitably affect the efficiency and accuracy of 3D reconstruction. To obtain realistic and cost-effective 3D models, weather data needs to be denoised and simplified prior to 3D reconstruction. There are two different ways of integrating cloud data and detailed cloud data from 3D laser scanners. To facilitate the full utilization of the data during 3D reconstruction, it is necessary to combine the two data into one joint [17].

**4.1.1 Point cloud data classification**

In the 3D reconstruction work, the process analysis and reconstruction methods of the feature content are directly related to the point cloud separation rules, so the point cloud distribution is also an important impact link in the 3D reconstruction work. The data cost after denoising and reduction is still huge. In order to improve the efficiency and automation of reconstruction, it is necessary to choose a distribution process suitable for the learning environment to improve the efficiency of 3D reconstruction [18]. According to the topographic and geographical features of the South-to-North Water Diversion Project, this paper divides it into two categories: three-dimensional features and rapid development of topography. First, separate the ground data from the data cloud data, and define the initial content of the data and its community scale, which is suitable for the area where the data cloud data area is located. space, and then disappears. Synonymous cloud data. The classification, content process analysis and construction process in 3D reconstruction are directly related to the content distribution in the cloud, so the weather distribution is also an important part of the 3D construction work. The data cost after denoising and reduction is still huge. In order to improve the efficiency and automation of reconstruction, an appropriate distribution needs to be selected for the learning environment to improve the efficiency of 3D reconstruction [19].

By the above method all point clouds are divided into two sets: terrain points and surface points. The terrain is mainly composed of roads and landforms, which will be used for terrain reconstruction; the point cloud data of the ground object points including buildings, signs and vegetation will be used to construct the ground object model, and the classification method of the decision tree will be used to classify the ground points and non-ground objects. The ground is further subdivided into: terrain, roads, buildings and vegetation [20]. Figure 5 shows the classification process:

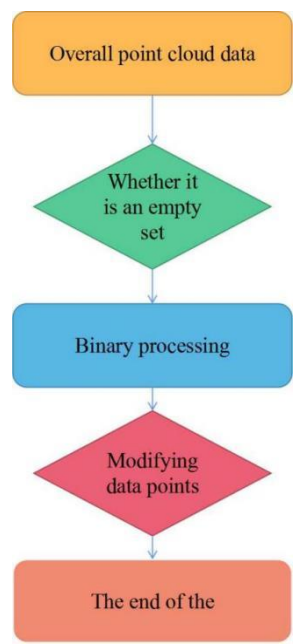

Figure 5 Classification flow chart

## 4.2 3D model reconstruction

In this paper, based on the classification results, the 3D model reconstruction of the entire scene is divided into landform reconstruction and feature reconstruction. The specific process is shown in Figure 6.

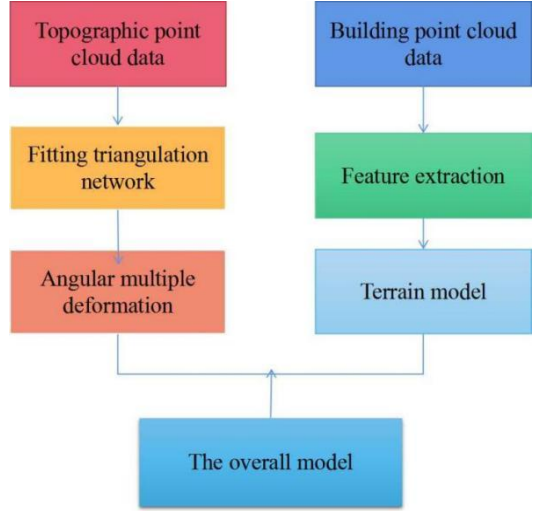

Figure 6 Flowchart of large-scale model reconstruction

## 4.3 Split wire reconstruction based on improved DPC algorithm.

The core of reconstructing the split conductor is to determine which split sub-conductor each point in the point cloud of the power line belongs to [21]. To this end, this paper firstly calculates the relative coordinates of the spatial points of the power line point cloud, and then proposes an improved DPC clustering algorithm based on relative coordinates, and applies it to the identification of the number of split conductors and the determination of the belonging and finally adopts the least squares method. Fit each splitter wire [22].

### 4.3.1 Calculation of relative coordinates of power line space points

There is no difference in the data whether each circuit is an insulated conductor or not. In this form, the center of the line segment obtained by balancing the parabola by the least squares method is then calculated as the relative sum of the element cloud terms of the relative velocity of the parabola. The special steps of the relational algorithm can be expressed as follows:

(1) Take any section of the power line point cloud data (this paper takes 10 m power line to ensure enough data), assuming that there are n sample points in this point cloud, and the coordinates of the point Pi are (xi, yi, zi), (i=2, 4, 8, ..., n);

(2) Convert the example point in the triangle to the example point in the two planes where the curve is located, then the sum of the points on both sides is (wi, zi), as shown in formula (1)

$$w_i = \sqrt{x_i^1 + y_i^1} \quad (1)$$

(3) The least squares method is used to fit n two-dimensional sample points (w, z; ), and the fitted parabolic equation is as shown in formula (2)

$$z = Aw^2 + B_w + c \quad (2)$$

## 4.4 Extraction of split sub-wires based on improved DPC algorithm

Because the point cloud space points belonging to different split sub-conductors are far apart, and the point cloud space points belonging to the same split sub-conductor are relatively close, that is, the relative spatial points on the same split sub-conductor are densely clustered, so , the point cloud extraction of split sub-wires can be realized by density clustering algorithm [23]. At present, the DPC algorithm, as a commonly used density clustering algorithm, can be used to quickly find the cluster centers of the dataset, and efficiently allocate sample points and eliminate outliers. However, this method also has the disadvantage that it requires human-computer interaction to make decisions and cannot automatically select cluster centers. For example, in the case of split wires, the traditional DPC algorithm cannot automatically determine the number of split sub-wires. Therefore, this paper proposes an improved DPC algorithm for automatic identification and extraction of split wires. The algorithm includes: normalization of decision parameters; drawing of decision diagram; discrimination of the number of split sub-wires. The specific algorithm is as follows [24].

Decision parameter normalization. The DPC algorithm includes two important decision parameters: the local density $\rho$ of sample points; the distance $\delta$ of sample points.

In this step, the relative coordinates of the spatial points of the power lines are input first, and the local density $\rho_j$ of each sample (spatial) point n (n=2, 4, 6, ., m) is calculated, and the calculation expression is as shown in Equation 3：

$$p = \sum_{j=i} \exp\left[\left(\frac{d_j}{d_e}\right)\right] \quad (3)$$

In the formula: dij is the Euclidean distance between sample points i and j; dc is the cut-off distance, which is set as the maximum diameter of the split sub-conductor in this paper, that is, dc = 0.05 m. In addition, find the sample point h with the largest $\rho$ value, and record its $\rho$ value as $\rho$max.

## 4.5 Performance comparison of different algorithms for the same model

This paper calculates the reconstruction results of different models, and selects the starting point cloud number, the number of triangle meshes, the Delaunay internalization period, and the total time as comparisons, as shown in Table 1. It can be seen that this form of algorithm fragmentation optimization With a long time, the resistance of Delaunay's internalization period is significantly reduced, mainly because the point cloud data must be traversed in the sharding optimization process. One cycle, in the triangulation process, the main triangulation is done in 2D space by a 2D mapping method, which greatly reduces the time of Delaunay internalization [25].

Table 1 3D reconstruction data of different models

| model name | Number of point clouds | Fragmentation time/ms | mapping time/ms | Optimization time/ms | Total time/ms |
|---|---|---|---|---|---|
| Bunny | 36952 | 7896 | 653 | 1456 | 1236 |
| Chari | 25986 | 6523 | 236 | 1236 | 9563 |
| Block | 12569 | 1569 | 698 | 6952 | 8956 |
| Dragon | 23394 | 2389 | 741 | 4569 | 7856 |

Table 1 compares the time consumption of this algorithm with Crust algorithm, Tight Cocone algorithm and DBRG algorithm. significantly faster than the Crust algorithm and the Tight Cocone algorithm, and can quickly reconstruct large-scale point cloud data.

Comparing different algorithms, it can be found that the time used by the algorithm in this paper is basically close to that of the DBRG algorithm, and the method adopted by this algorithm to solve the restoration problem and the algorithm in this paper both include the following steps: expansion of the region growth method and Delaunay gridding. The difference is that the method in this paper performs Delaunay internalization in the 2D plane. After the Delaunay internalization of the grouped point cloud, it needs to be spliced and optimized to consume part of the time, while the DBRG algorithm is to perform Delaunay triangulation in space. Although it takes a long time to carry out Delaunay internalization in 3D space, the direct application of the regional growth rule to the triangulated patch in the space eliminates the splicing link and reduces the area expansion time, so the time spent In terms of time, it is slightly better than the algorithm in this paper. The Crust algorithm and the Cocone algorithm are based on "sculpting" algorithms, which need to peel off the tetrahedral cells in the entire space layer by layer to obtain the original surface of the object to be built, thus greatly increasing the time consumption in this regard.

**5 Conclusion**

This chapter introduces the related research status of point cloud 3D reconstruction, and divides point cloud 3D reconstruction into two categories according to the requirements for sampling point sets: interpolation method and approximation method. The former type has stricter requirements on sampling point set, and the restored model The surface will pass through all the original data points, which requires the parametric expression of the topology. Therefore, the topology is often the most accurate, but it is difficult to realize, and it is often suitable for the restoration of some small models with known topology. Its representative methods include the RBF interpolation function reduction method, and the surface reconstruction method of different topological B-spline models. The approximation method requires less sampling conditions, but the final sampling result is often only a good approximation of the original surface, and cannot fully guarantee the correctness of the original effect. The incremental surface reconstruction method has been used in recent years because of the The reconstruction effect is good, the thinking is relatively simple and the operation is relatively fast, which has been widely cited. The algorithm proposed in this paper also utilizes this method to shard point cloud data.

Photogrammetry and Remote Sensing, 79(5), 29-43.